\PassOptionsToPackage{dvipsnames}{xcolor}
\documentclass[sigconf]{acmart}
\usepackage{subcaption}
\usepackage{booktabs} 
\usepackage{multirow} 
\usepackage{newunicodechar}
\newunicodechar{（}{(}
\newunicodechar{）}{)}
\usepackage{siunitx} 
\usepackage{graphicx} 
\usepackage{amsmath} 
\usepackage{makecell} 
\usepackage{booktabs}
\usepackage{subcaption}
\usepackage{bbding}
\usepackage{float}
\usepackage{booktabs}
\usepackage{caption}
\AtBeginDocument{%
  }

\copyrightyear{2025}
\acmYear{2025}
\setcopyright{cc}
\setcctype{by}
\acmConference[MMAsia '25]{ACM Multimedia Asia}{December 9--12, 2025}{Kuala Lumpur, Malaysia}
\acmBooktitle{ACM Multimedia Asia (MMAsia '25), December 9--12, 2025, Kuala Lumpur, Malaysia}\acmDOI{10.1145/3743093.3770987}
\acmISBN{979-8-4007-2005-5/2025/12}

\begin{document}

\title{DualCap: Enhancing Lightweight Image Captioning via \\ Dual Retrieval with Similar Scenes Visual Prompts}



\author{Binbin Li}
\affiliation{%
  \institution{Institute of Information Engineering, Chinese Academy of Sciences}
  \city{}
  \country{}}
\affiliation{
  \institution{School of Cyber Security, University of Chinese Academy of Sciences}
  \city{Beijing}
  \country{China}}
\email{libinbin@iie.ac.cn}

\author{Guimiao Yang}
\affiliation{%
  \institution{Institute of Information Engineering, Chinese Academy of Sciences}
  \city{}
  \country{}}
\affiliation{
  \institution{School of Cyber Security, University of Chinese Academy of Sciences}
  \city{Beijing}
  \country{China}}
\email{yangguimiao@iie.ac.cn}

\author{Zisen Qi}
\authornote{Corresponding author: Zisen Qi.}
\affiliation{
  \institution{Institute of Information Engineering, Chinese Academy of Sciences}
  \city{Beijing}
  \country{China}}
\email{qizisen@iie.ac.cn}

\author{Haiping Wang}
\affiliation{
  \institution{Institute of Information Engineering, Chinese Academy of Sciences}
  \city{Beijing}
  \country{China}}
\email{wanghaiping@iie.ac.cn}

\author{Yu Ding}
\affiliation{
  \institution{Institute of Information Engineering, Chinese Academy of Sciences}
  \city{Beijing}
  \country{China}}
\email{dingyu@iie.ac.cn}



\renewcommand\footnotetextcopyrightpermission[1]{}

\begin{abstract}
  Recent lightweight retrieval-augmented image caption models often utilize retrieved data solely as text prompts, thereby creating a semantic gap by leaving the original visual features unenhanced, particularly for object details or complex scenes. To address this limitation, we propose $DualCap$, a novel approach that enriches the visual representation by generating a visual prompt from retrieved similar images. Our model employs a dual retrieval mechanism, using standard image-to-text retrieval for text prompts and a novel image-to-image retrieval to source visually analogous scenes. Specifically, salient keywords and phrases are derived from the captions of visually similar scenes to capture key objects and similar details. These textual features are then encoded and integrated with the original image features through a lightweight, trainable feature fusion network. 
  Extensive experiments demonstrate that our method achieves competitive performance while requiring fewer trainable parameters compared to previous visual-prompting captioning approaches.
   
\end{abstract}






\maketitle

\section{Introduction}

\begin{figure}[t!]
    \vspace*{0.35cm} 
    \hspace*{\dimexpr\linewidth-0.99\linewidth} 
    \begin{minipage}{0.85\linewidth} 
        \begin{subfigure}[t]{\linewidth} 
            \includegraphics[width=\linewidth, keepaspectratio]{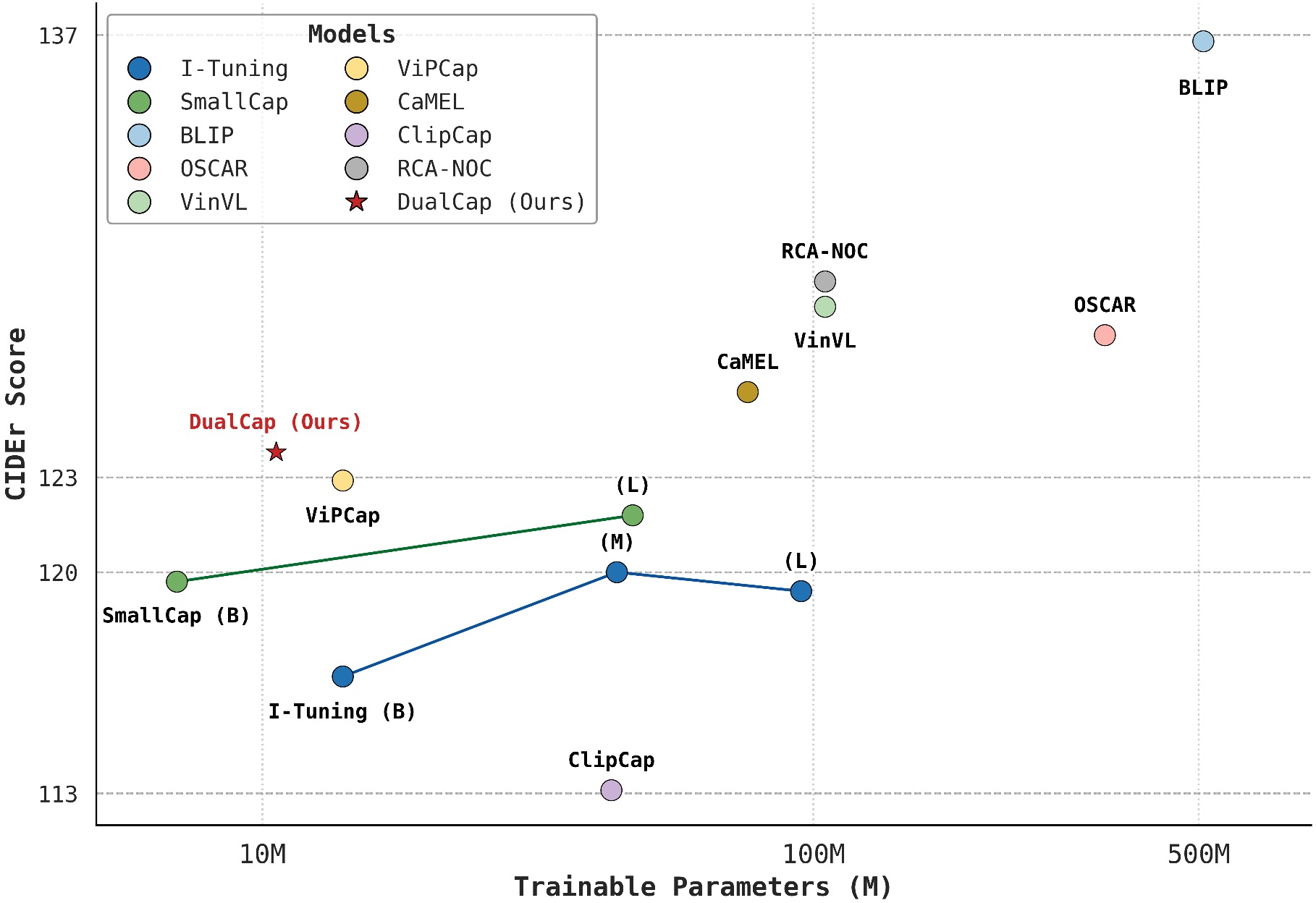}
            \caption{The performance-complexity trade-off across model scales, quantifying both parameter counts and CIDEr evaluation scores (B: Base, M: Medium, L: Large).}
            \label{fig:model1a}
        \end{subfigure}
        
        \vspace{0.3cm} 
        
        \begin{subfigure}[t]{\linewidth} 
            \includegraphics[width=\linewidth, keepaspectratio]{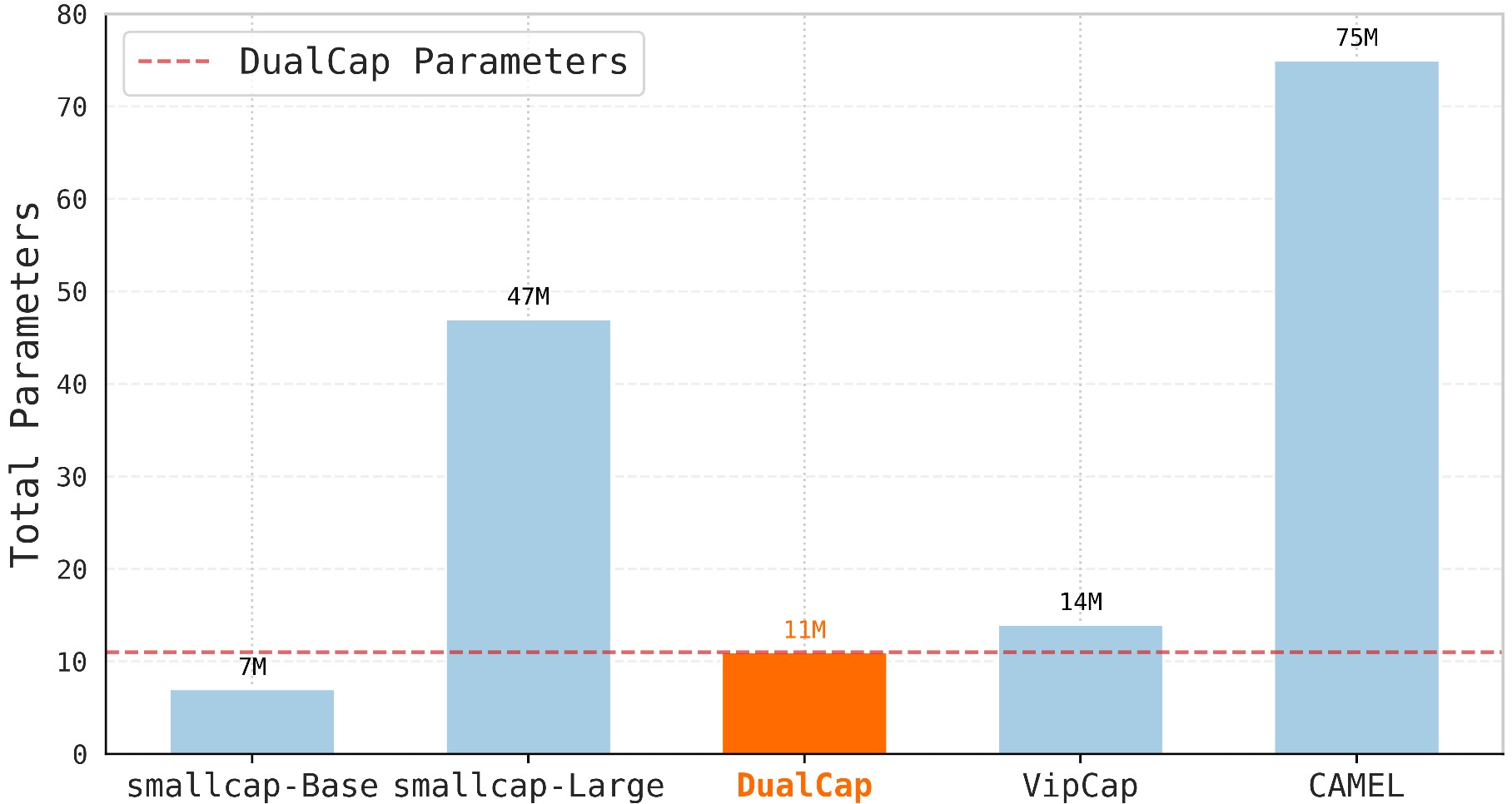}
            \caption{Comparison of total parameters between DualCap (Ours) and
other lightweight models.}
            \label{fig:model1b}
        \end{subfigure}
        \label{fig:model1}
    \end{minipage}
    \caption{(a) DualCap shows the best efficiency among the lightweight captioning models. (b) Comparison of total model parameters, the DualCap requires only 11M parameters, demonstrating its lightweight design.}
    \Description{Performance comparison of DualCap with other lightweight models, showing efficiency and parameter count trade-offs.}
\end{figure}

The state-of-the-art in image captioning is currently defined by the advancement of large-scale vision-language (V\&L) models, such as SimVLM\cite{simvlm}, PaLi\cite{pali}, and REVEAL\cite{reveal}, which are trained on ever-expanding datasets. This trend of scaling up, however, has introduced significant challenges. These advanced multimodal models require a vast number of parameters, frequently in the billions, which results in high computational demands and expensive dataset construction costs. This issue is particularly exacerbated during the pre-training and fine-tuning stages for downstream tasks, making it difficult to adapt and deploy numerous model versions for different visual domains or specific real-world applications.

To enhance training efficiency, some models have focused on lightweight architectures that freeze large pre-trained models and only train a small set of parameters. Approaches like ClipCap\cite{clipcap}, I-Tuning\cite{ituning}, CAMEL\cite{camel}, and BLIP\cite{blip1} have introduced mapping networks or learnable tokens to bridge the modality gap. Nevertheless, even these more efficient methods can be resource-intensive; for instance, BLIP-2\cite{blip2} still requires over 1B parameters, while others like EVCap\cite{evcap} depend on high-performance models with over 5B total parameters, limiting their practicality and accessibility. As illustrated in Figure \ref{fig:model1b}, the parameter scale of these models dwarfs that of more efficient approaches, motivating a critical shift towards Retrieval-Augmented Generation, which leverages external knowledge to enhance caption quality while minimizing the number of trainable parameters.

Recent efforts like SmallCap\cite{ramos2022smallcap}, EXTRA\cite{extra}, and LMCap\cite{lmcap} leverage external knowledge by retrieving semantically similar texts to use as prompts to reduce computational costs. This method not only reduces the training burden but also enables training-free domain adaptation by simply replacing the external datastore. However, these approaches present a key limitation: by using the retrieved content solely as text prompts, the visual information remains static, relying only on the initial CLIP vision encoder output. This creates a semantic gap where the rich descriptions within the text prompt are not utilized to enhance the model's visual understanding. 

To address this, recent work has explored using retrieved text to create visual prompts, which directly enhance the visual representations before the decoder processes them. A pioneering example, ViPCap\cite{vipcap}, generates a visual prompt by modeling the embedding of a retrieved caption as a learnable Gaussian distribution\cite{vae} and fusing sampled semantic features with the original image features. 
While this demonstrates the importance of enhancing the visual stream, the approach depends on a single retrieval source and incorporates the entire retrieved caption, potentially introducing noise or irrelevant information.

From the captions of these visually similar scenes, DualCap employs a more focused strategy. Instead of using entire sentences, it extracts salient scene-keywords and phrases that capture the core objects, actions, and attributes. These targeted textual features are then encoded and integrated with the original patch-level visual features via a lightweight, trainable Feature Fusion Network called SFN\cite{albef}. This process creates a refined visual representation that is enriched with relevant, contextually-grounded semantic details.

Our main contributions can be summarized as follows:
\begin{itemize}
    \item We propose DualCap, a lightweight image captioning framework featuring a novel dual retrieval mechanism that separates the sourcing of textual and visual prompts.
    \item This work presents a method for generating potent visual prompts based on scene-keywords extracted from visually similar images, offering a more targeted and less noisy alternative to using full captions.
    \item Extensive experiments demonstrate that DualCap achieves state-of-the-art performance among lightweight models on the COCO, Flickr30k, and NoCaps benchmarks, outperforming prior methods with fewer trainable parameters and showing superior generalization to novel objects.
\end{itemize}

\section{Related Work}

\subsection{Retrieval-based image captioning}
The substantial computational and data demands of large-scale vision-language models have prompted the development of more efficient alternatives. To enhance vision-language integration without increasing parameter counts, Retrieval-augmented generation has been effectively integrated. Methods like SmallCap (Ramos et al. 2023), LMCap \cite{lmcap}, EVCap \cite{evcap}, and MeaCap \cite{MeaCap} use text-based datasets and store captions in a datastore. SmallCap\cite{ramos2022smallcap}, for instance, generates captions using an input image and retrieved text from a datastore, requiring only 7 million trainable parameters to enable efficient training. However, its visual understanding capabilities remain limited. To address this, our method enhances performance by generating visual prompts from a dual retrieval mechanism containing detailed image descriptions.

\subsection{Visual prompt with textual information}
Prompt-based learning~\cite{prefixtuning,power} has emerged as a powerful, parameter-efficient paradigm for adapting large-scale pretrained models to downstream tasks across both vision-only and V$\&$L domains~\cite{vpt,dualprompt}. The visual prompts are inserted into a ViT or task-specific patches to steer model behavior. However, a fundamental semantic gap persists between the static visual features of an image and the rich context available in retrieved texts. A pioneering work by ViPCap \cite{vipcap} sought to bridge this gap by introducing an innovative retrieval text-based visual prompt.  Instead of merely prompting the text decoder, ViPCap leverages the embeddings of retrieved captions to generate a dynamic visual prompt that is fused with the original image features. Although this approach demonstrated considerable efficacy, its reliance on sampling from a randomized Gaussian distribution can introduce significant noise, potentially degrading the prompt's quality and stability. To overcome this limitation, we propose a novel visual prompt constructed from keywords that are semantically grounded in similar visual scenes, thereby ensuring a more stable and contextually relevant prompt generation process.

\subsection{Exploiting localized features and keywords}
The advanced lightweight captioning models almost rely on global image features, which may overlook fine-grained details and the relationships between objects. To generate more detailed and accurate descriptions, researchers \cite{enhance-ic} have explored using more localized visual information and specific keywords. By extracting keywords, the model can concentrate on essential objects and reduce the noise often present in full-sentence captions. This strategy of using targeted textual elements provides a strong motivation for our approach of generating visual prompts from specific scene-keywords.

\begin{figure*}[t]
\centering
\includegraphics[width=\textwidth]{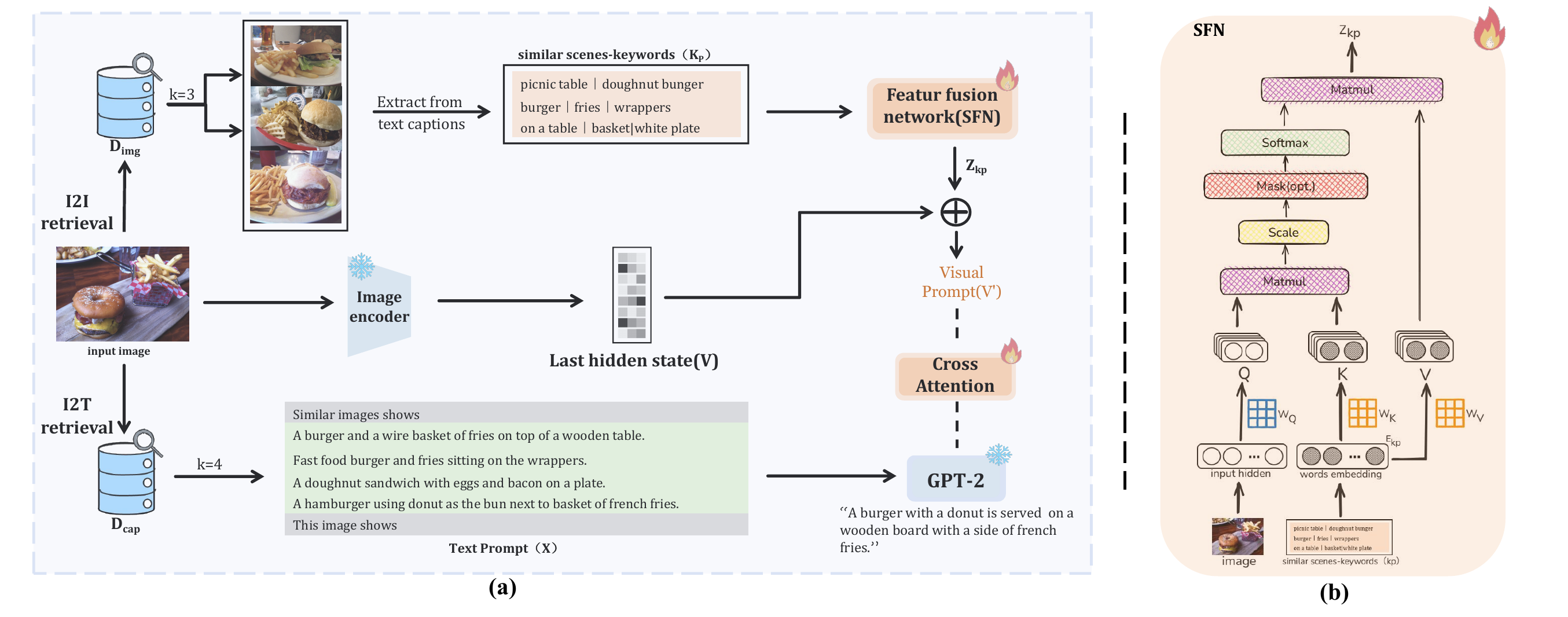} 
\caption{(a) DualCap generates detailed captions by feeding a GPT-2 decoder two parallel prompts: a text prompt X from retrieved captions (I2T path), and an enhanced visual representation V'. The latter is created by using the SFN to generate a visual prompt $Z_{kp} $from keywords of similar scenes (I2I path), which is then added to the original image features V. (b) The architecture of the SFN, it employs a cross-attention mechanism where image patch features V act as the Query to attend to the semantic information from scene-keyword embeddings ($E_{kp}$), which serve as Key and Value.}
\Description{The DualCap framework diagram showing two parallel retrieval paths: Image-to-Image (I2I) and Image-to-Text (I2T), which generate a visual prompt and a text prompt respectively, to produce a detailed image caption using GPT-2.}
\label{dualcap}
\end{figure*}

\section{Proposed Method}

Our model, DualCap, is designed to address a critical limitation in existing lightweight image captioning frameworks: the difficulty in generating descriptions rich with fine-grained visual details. While models like SmallCap\cite{ramos2022smallcap} are parameter-efficient, their reliance on a single stream of semantic retrieval often overlooks specific visual nuances. To overcome this, DualCap introduces a novel architecture built upon a frozen CLIP ViT-B/32 encoder $E_{\mathbf{v}}$\cite{clip} and a GPT-2 decoder $E_{\mathbf{D}}$\cite{gpt2}. The core of our innovation, as illustrated in Figure \ref{dualcap}, is a dual-stream retrieval mechanism that strategically decouples the acquisition of broad semantic context from that of specific visual evidence. This is coupled with a scene-aware, keyword-driven visual prompting module, which refines the image representation itself before decoding.

The framework operates by generating two distinct inputs for the decoder: a text prompt $\mathbf{X}$, derived from semantically similar captions, which sets the general context; and a refined visual feature representation $\mathbf{V^{'}}$, which has been enriched with precise details from visually analogous scenes. This dual-input strategy allows the model to synergistically leverage both high-level semantic guidance and low-level visual specificity, leading to more accurate and detailed captions.

\subsection{Dual retrieval mechanism}
DualCap employs two parallel retrieval streams to gather distinct types of contextual information for the captioning task. Let $\mathcal{D}_{img}$ be the datastore of pre-computed training image features and $\mathcal{D}_{cap}$ be the datastore of corresponding caption features. 

\noindent\textbf{Image-to-Text (I2T) Retrieval for Text Prompt.} Following the successful approach of SmallCap, we use a standard image-to-text retrieval path to construct a textual prompt. For a given input image $I$, we use its global CLIP feature $\mathbf{v}_I=E_{\mathbf{v}}(I)$ to query the caption datastore $\mathcal{D}_{cap}$. We retrieve the top-k most semantically similar captions based on cosine similarity, forming the set $C_{I2T} = \{c_{1},...,c_{k}\}$. These captions are then inserted into a hard prompt template $\mathbf{X}$\cite{kgcoop}: \texttt{Similar images show} \{\texttt{caption$_1$}\}\texttt{...}\{\texttt{caption$_k$\}}. \texttt{This image shows \_\_\_}. Then, this prompt initializes the GPT-2 decoder, providing contextual examples that guide the overall theme and style of the generated caption. 

\noindent\textbf{Image-to-Image (I2I) Retrieval for Visual Prompt.} This is the first core component of our novel design. Unlike the I2T path, the I2I path aims to find images that are visually analogous to the input. For a given input image $I$ with its feature vector $\mathbf{v}_I$, we query the image feature datastore $\mathcal{D}_{img}$ to retrieve the top-M most visually similar images. This retrieval process is formally defined as:
\begin{equation}
    \{\mathbf{I}_1, ..., \mathbf{I}_M\} = \underset{\mathbf{I}_j \in \mathcal{D}_{img}}{\text{arg top-M}} \left( \frac{\mathbf{v}_I^T \mathbf{v}_j}{\|\mathbf{v}_I\| \|\mathbf{v}_j\|} \right)
\end{equation}
where the operation retrieves the M images $I_j$ from the datastore that maximize the cosine similarity with the input feature $\mathbf{v}_I$. We hypothesize that the captions associated with this retrieved set, $C_{I2I} = \{ cap(I_1),...,cap(I_M)\}$, contain descriptions of objects, attributes, and actions that are highly relevant to the visual content of the input image. This set forms a high-fidelity source for generating our visual prompt.

\subsection{Similar Scene-Keywords Visual Prompting}

\noindent\textbf{Scene-Keyword Extraction and Embedding.} Instead of using the full sentences from the retrieved captions $C_{I2I}$, we introduce a more precise mechanism to distill this information and inject it directly into the visual features.
The goal is to extract a set of precise, high-signal keywords, denoted as $K_{p}$. To distill these core concepts, we process the retrieved captions $C_{I2I}$ using a standard NLP toolkit NLTK\cite{nltk}. The process involves several steps, including part-of-speech (POS) tagging and rule-based chunking. This entire extraction procedure can be formally defined as:
\begin{equation}
    K_p = \bigcup_{c_i \in C_{I2I}} \left\{ w \mid w \in \text{Chunk}(c_i) \land \text{POS}(w) \in \mathcal{T}_{\text{kp}} \right\}
\end{equation} For each caption $c_i$ in the retrieved set $C_{I2I}$, we extract candidate words and phrases $w$ using a chunking function($\text{Chunk}$). We then filter these candidates, retaining only those whose part-of-speech tag ($POS(w)$) belongs to a predefined set of valuable tags $\mathcal{T}_{\text{kp}}$. The union operation $\bigcup$ ensures that the final set contains $K_{p}$ only unique elements. This procedure effectively filters out grammatical boilerplate and irrelevant contextual details, producing a concentrated set of semantic anchors for visual grounding.

These scene-keywords capture the essential semantic essence of the similar scenes. Each keyword $kp_i$ is then encoded using the frozen CLIP text encoder $\mathbf{E_T}$, yielding a set of keyword embeddings $E_{kp} = \{e_1,...,e_n\}$, where $e_i = \mathbf{E_T}(kp_i)$. The resulting tensor $E_{kp} \in R^{P\times D_{text}}$ represents the distilled semantic information ready for fusion. 

\noindent\textbf{Feature Fusion Network(SFN).} The Feature Fusion Network (SFN) is a trainable, lightweight, and single-layer Transformer\cite{transformer} designed to fuse the semantic keyword information with the patch-level visual features. This network takes the original image patch features $\mathbf{V} \in R^{N\times D_{vision}}$ and the keyword embeddings $E_{kp} \in R^{P\times D_{textn}}$ as inputs. It employs a cross-attention mechanism where the visual features act as the \textbf{Query}, while the keyword embeddings serve as both \textbf{Key} and \textbf{Value}. This design allows each image patch to "attend to" the most relevant keywords, thereby grounding semantic concepts to specific visual regions. The modeling process is as follows. First, the inputs are mapped to a common attention space via trainable projection matrices $W_Q \in R^{D_{vision} \times d_k}$,$W_K \in R^{D_{text} \times d_k}$, and $W_V \in R^{D_{text} \times d_k}$:
\begin{equation}
    \mathbf{Q}=VW_Q,\mathbf{K}=E_{kp}W_K,\mathbf{Val}=E_{kp}W_V
\end{equation} Next, the attention output is computed using the scaled dot-product attention formula, where $d_k$ is the attention dimension:
\begin{equation}
    Attention(\mathbf{Q},\mathbf{K},\mathbf{Val}) = softmax(\frac{\mathbf{Q}\mathbf{K}^{T}}{\sqrt{d_{k}}})\mathbf{Val}
\end{equation} The final output of SFN is a visual prompt $Z_{kp} \in R^{N \times D_{vision}}$, which can be interpreted as a semantic adjustment vector that enhances each patch's representation with the weighted semantics of the most relevant scene keywords.\\ \textbf{Caption Generation and Training.} The final refined visual representation $\mathbf{V^{'}}$ is created by adding the visual prompt to the original features via a residual connection:
\begin{equation}
    \mathbf{V^{'}}=\mathbf{V}+\mathbf{Z_{kp}}
\end{equation} This enhanced representation $\mathbf{V^{'}}$, which now contains both the raw visual data and targeted semantic cues, is passed to the cross-attention layers of the GPT-2 decoder. The decoder then generates the final caption token by token, conditioned on both the text prompt $\mathbf{X}$ and the refined visual features $\mathbf{V^{'}}$. \\ The entire model is trained end-to-end by minimizing the standard cross-entropy loss $\mathcal{L_{\theta}}$ for a target caption $Y = \{y_{1},...,y_{T}\}$:
\begin{equation}
    \mathcal{L} _{\theta} =-\sum_{t=1}^{T} logP_{\theta }(y_{t}|y_{<t},X,V^{'})
\end{equation} This objective function trains the model to predict the next word in the sequence, given the preceding ground-truth words, the high-level text prompt, and our detail-enriched visual features.

\section{Experiments}

\begin{table*}[ht]
\centering
\caption{Comparison with large pre-trained and lightweight models with existing methods on the COCO test, Flickr30k test, and NoCaps val set. CIDEr score is used for NoCaps evaluation. Our method shows the competitive performance in most metrics.}
\resizebox{\textwidth}{!}{
\begin{tabular}{@{}l|c|cccc|cc|c|c|c|c@{}}
\toprule
\multirow{2}{*}{\textbf{Method}} & \multirow{2}{*}{\makecell{\textbf{Training}\\\textbf{Param $\boldsymbol{\theta}$}}} & \multicolumn{4}{c|}{\makecell{\textbf{COCO}\\\textbf{Test}}} & \multicolumn{2}{c|}{\makecell{\textbf{Flickr30k}\\\textbf{Test}}} & \multicolumn{4}{c}{\makecell{\textbf{NoCaps}\\\textbf{Val}}} \\
& & B@4 & M & C & S & C & S & In & Near & Out & Entire \\
\midrule
\multicolumn{12}{@{}l}{\textbf{Large scale training models}} \\
OSCAR$_{\text{Large}}$ (2020) & 338M & 37.4 & 30.7 & 127.8 & 23.5 & - & - & 78.8 & 78.9 & 77.4 & 78.6 \\
Lemon$_{\text{HUGE}}$ (2022) & 675M & 41.5 & 30.8 & 139.1 & 24.1 & - & - & 118.0 & 116.3 & 120.2 & 117.3 \\
SimVLM$_{\text{Huge}}$ (2022a) & 632M & 40.6 & 33.7 & 143.3 & 25.4 & - & - & 113.7 & 110.9 & 115.2 & 112.2 \\
BLIP2$_{\text{ViT-g OPT}_{2.7\text{B}}}$ (2023a) & 1.1B & 43.7 & - & 145.8 & - & - & - & 123.0 & 117.8 & 123.4 & 119.7 \\
CogVLM$_{\text{2.7B}}$ (2024) & 1.5B & - & - & 148.7 & - & 94.9 & - & - & - & 132.6 & 128.3 \\
PaLI$_{\text{mT5-XXL}}$ (2023) & 1.6B & - & - & 149.1 & - & - & - & - & - & - & 127.0 \\
\midrule
\multicolumn{12}{@{}l}{\textbf{Lightweight models}} \\
CaMEL (2022) & 76M & \textbf{39.1} & \textbf{29.4} & \textbf{125.7} & \textbf{22.2} & - & - & - & - & - & - \\
I-Tuning$_{\text{Medium}}$ (2023) & 44M & 35.5 & \underline{28.8} & 120.0 & 22.0 & \textbf{72.3} & \textbf{19.0} & 89.6 & 77.4 & 58.8 & 75.4 \\
ClipCap (2021) & 43M & 33.5 & 27.5 & 113.1 & 21.1 & - & - & 84.9 & 66.8 & 49.1 & 65.8 \\
I-Tuning$_{\text{Base}}$ (2023) & 14M & 34.8 & 28.3 & 116.7 & 21.8 & 61.5 & 16.9 & 83.9 & 70.3 & 48.1 & 67.8 \\
SmallCap (2023) & 7M & 37.0 & 27.9 & 119.7 & 21.3 & 60.6 & - & 87.6 & 78.6 & 68.9 & 77.9 \\
SmallCap$_{\text{id=16, Large}}$ (2023) & 47M & 37.2 & 28.3 & 121.8 & 21.5 & - & - & - & - & - & - \\
VipCap (2025) & 14M & 37.7 & 28.6 & 122.9 & 21.9 & 66.8 & 17.2 & 93.8 & 81.6 & 71.5 & 81.3 \\
\textbf{DualCap (Ours)} & 11M & \underline{37.9} & 28.6 & \underline{123.6} & 22.0 & \underline{68.5} & \underline{18.3} & \textbf{94.3} & \textbf{82.7} & \textbf{72.2} & \textbf{81.9} \\
\bottomrule
\end{tabular}%
} 
\label{tab:model_comparison_final_v2} 
\end{table*}

\subsection{Experimental Setup}

\noindent\textbf{Training dataset.}Our experiments are conducted on three standard image captioning benchmarks: COCO~\cite{coco}, NoCaps~\cite{nocaps}, and Flickr30k~\cite{plummer2016flickr30k}. Following common practice, we use the Karpathy splits~\cite{karpathysplits} for training, validation, and testing on COCO and Flickr30k.
Our model is trained exclusively on the COCO training set. We report our primary results on the COCO Karpathy test set, while the Flickr30k test set and the NoCaps validation set are used to assess the model's transferability and generalization capabilities, respectively.

\noindent\textbf{Training setup.} DualCap includes a vision encoder CLIP ViT-B/32~\cite{vit} and a language decoder GPT-2$_{\text{Base}}$~\cite{huggingface}, both of which are frozen during training.We train only the core trainable components: (1) the cross-attention layers within the GPT-2 decoder, following SmallCap~\cite{ramos2022smallcap}, and (2) our newly introduced Feature Fusion Network (SFN). The SFN and the cross-attention layers are implemented as single-layer blocks with a 12-head attention mechanism, and we scale the attention dimension to 32 to reduce computational cost.DualCap consists of 11M trainable parameters and is trained for 10 epochs on a single NVIDIA A100 GPU with a batch size of 64.The model is optimized using the AdamW optimizer~\cite{adamw} with an initial learning rate of $1 \times 10^{-4}$ and a weight decay of 0.05. We select the checkpoint with the best CIDEr score on the validation set.

During training, the model is prompted with inputs from our dual retrieval mechanism: we retrieve the $top - 4$ most semantically similar captions for the text prompt, and for the visual prompt, we retrieve the top-3 most visually similar images, from which we extract up to $P = 12$ scene-keywords.The choice of k=4 for the text prompt was determined through preliminary experiments\cite{ramos2022smallcap}, offering a robust contextual basis without introducing excessive noise from less relevant captions. Both retrieval streams use pre-computed CLIP-ResNet-50x64~\cite{clip} representations, indexed with FAISS~\cite{faiss} for efficient searching. During inference, we generate captions using beam search decoding with a beam size of 3 on the COCO test set. All inference time measurements were conducted on a single NVIDIA A100 GPU. Caption quality is evaluated using BLEU@4~\cite{bleu}, METEOR~\cite{meteor}, CIDEr~\cite{cider}, and SPICE~\cite{spice}.

\subsection{Main Results}

\noindent\textbf{In-Domain Benchmarks:} As shown in Table \ref{tab:model_comparison_final_v2}, we evaluate DualCap on the COCO, Flickr30k. With only 11M trainable parameters, our model achieves a CIDEr score of 123.6 and a B@4 score of 37.9 on the COCO dataset. Compared to the SmallCap baseline, our model achieves a significant 3.26\% improvement in CIDEr score, increasing it from 119.7. Furthermore, this result outperforms ViPCap's 122.9 CIDEr, even though ViPCap utilizes more trainable parameters at 14M. This highlights the effectiveness of our dual retrieval and scene-keyword fusion mechanism in generating more accurate and contextually relevant captions for in-domain images.

\noindent\textbf{Cross-Domain Performance:} The advantages of DualCap are most pronounced in cross-domain and generalization settings, where it sets a new SOTA for lightweight models. On the NoCaps validation set, DualCap achieves an overall CIDEr score of 81.9. This not only surpasses similarly sized competitors like ViPCap at 81.3, but also significantly outperforms much larger models like the 44M-parameter I-Tuning$_{Medium}$, which scores only 75.4. The model's strength is particularly evident in its 72.2 CIDEr score on the demanding out-of-domain split, a significant improvement over ViPCap's 71.5 and SmallCap's 68.9. The Image-to-Image retrieval path and Semantic Fusion Network(SFN) in concert to source and integrate semantic information, granting our model enhanced generalization capabilities that exceed even those of larger architectures.

\noindent\textbf{Inference Time:} As detailed in Table \ref{inference}, we assess the inference efficiency of DualCap against other lightweight models. Compared to high-performance models like CaMEL, DualCap is substantially more efficient, requiring only 11M trainable parameters versus CaMEL's 76M and reducing inference time from 0.56 to 0.42 seconds per image while maintaining competitive performance. Although our dual-retrieval mechanism introduces a modest computational overhead relative to the SmallCap baseline, raising inference time from 0.25 seconds, this is justified by a significant 3.9-point improvement in CIDEr score. This result demonstrates that DualCap offers a superior trade-off between descriptive accuracy and computational efficiency, achieving near state-of-the-art results without the heavy latency of larger models.

\begin{table}[h]
\centering
\caption{Comparison of different models on the COCO test set of inference time, trainable parameters($\theta$), and CIDEr score.}
\begin{tabular}{l|cccc}
\toprule
\textbf{Model} & \textbf{Inference Time(s)} & \textbf{$\theta$(M)} & \textbf{CIDEr} \\
\midrule
ClipCap & 0.16 & 43 & 111.7 \\
CaMEL & 0.56 & 76 & 125.7 \\
SmallCap & 0.25 & 7 & 119.7 \\
\midrule
\textbf{DualCap} & \textbf{0.42} & \textbf{11} & \textbf{123.6} \\
\bottomrule
\end{tabular}
\label{inference}
\end{table}

\subsection{Ablation Studies}
To comprehensively analyze the contributions of each component within our DualCap framework, we conduct detailed ablation experiments on the COCO dataset, including the Dual-retrieval mechanism, Semantic Fusion Network, and the various decoders.

\noindent\textbf{Analysis of Core Components.} We conducted a comprehensive ablation study to validate the contributions of our core components, with results summarized in Table \ref{core}. The experiments validate our hypothesis to decouple the retrieval of textual and visual evidence. The full model, which uses the Image-to-Image (I2I) stream to inform the SFN, outperforms an alternative where the SFN relies only on the text stream, achieving a CIDEr score of 123.6 versus 122.9. This confirms that sourcing visual keywords separately enhances performance, and the final model's top score highlights the effective synergy between all three components.

\begin{table}[h]
\centering
\caption{Incremental contribution of each component on the COCO test set. Each proposed module provides a distinct and significant improvement.}
\begin{tabular}{@{}lcc|cccc@{}}
\toprule
\textbf{I2T} & \textbf{I2I}  & \textbf{SFN} & \textbf{B@4} & \textbf{M} & \textbf{C} & \textbf{S}\\
\midrule
\Checkmark & \XSolidBrush & \XSolidBrush & 36.8 & 27.8 & 119.7 & 21.3 \\
\XSolidBrush & \Checkmark & \XSolidBrush & 34.9 & 26.7 & 113.5 & 20.6 \\
\midrule
\XSolidBrush & \Checkmark & \Checkmark & 35.2 & 28.3 & 116.4 & 21.2 \\
\Checkmark & \Checkmark & \XSolidBrush & 37.0 & 28.1 & 120.0 & 21.5 \\
\Checkmark & \XSolidBrush & \Checkmark & 37.7 & 28.5 & 122.9 & 21.8 \\
\midrule
\Checkmark & \Checkmark & \Checkmark & \textbf{37.9} & \textbf{28.6} & \textbf{123.6} & \textbf{22.0} \\
\bottomrule
\end{tabular}
\label{core}
\end{table}

\noindent\textbf{The Feature Fusion Network Design.} Our attention-based SFN significantly outperforms simpler fusion alternatives, as demonstrated in Table \ref{features}. While basic methods like direct summation, concatenation, or even the addition of an MLP layer\cite{mlp} yield only limited gains, our approach proves superior. Notably, to preserve the lightweight nature and training efficiency of our framework, this entire process is conducted with a frozen CLIP text encoder.

\noindent\textbf{Alternative Decoders.}
To ensure the benefits of our dual-stream architecture are not tied to a specific language decoder, we replaced the GPT-2 decoder with two different pre-trained models: OPT-125M\cite{opt} and the multilingual XGLM-564M\cite{xglm}. As shown in Table \ref{decoders}, while the baseline performance varies with each decoder, the performance gain from applying the DualCap architecture remains consistently high. This confirms that our framework's architectural benefits are robust and decoder-agnostic.


\begin{table}[!htbp]
\centering
\caption{Comparison of different feature fusion methods on the COCO test set.}
\setlength{\abovetopsep}{2pt} 
\setlength{\belowbottomsep}{6pt} 
\begin{tabular}{l|cccc}
\toprule
\textbf{Fusion Method} & \textbf{B@4} & \textbf{CIDEr} \\
\midrule
Sum & 36.6 & 120.5 \\
Concat & 37.2 & 121.4 \\
Concat + MLP & 37.6 & 121.9 \\
\midrule
\textbf{Ours (SFN+Cross-Attention)} & \textbf{37.9} & \textbf{123.6} \\
\bottomrule
\end{tabular}
\vspace{-2.0em}
\label{features}
\end{table}

\begin{table}[!htbp]
\centering
\caption{Model-agnostic evaluation with different decoders on the COCO test set using the Visual Prompt and the Text Prompt.}
\setlength{\abovetopsep}{2pt}
\setlength{\belowbottomsep}{6pt}
\begin{tabular}{llccc}
\toprule
\textbf{Method} & \textbf{Dec.} & \textbf{I2T} & \textbf{I2I} & \textbf{CIDEr} \\
\midrule
SmallCap  & GPT-2 (Base) & \Checkmark & \XSolidBrush  & 119.7 \\
\textbf{DualCap}  & \textbf{GPT-2 (Base)} & \Checkmark & \Checkmark & \textbf{123.6}({\color{ForestGreen}\textbf{3.9$\uparrow$}})\\
\midrule
SmallCap  & OPT-125M & \Checkmark & \XSolidBrush & 120.5 \\
\textbf{DualCap}  & \textbf{OPT-125M} & \Checkmark & \Checkmark & \textbf{124.2}({\color{ForestGreen}\textbf{3.7$\uparrow$}}) \\
\midrule
SmallCap  & XGLM & \Checkmark & \XSolidBrush & 118.9 \\
\textbf{DualCap}  & \textbf{XGLM} & \Checkmark & \Checkmark & \textbf{123.0}({\color{ForestGreen}\textbf{4.1$\uparrow$}}) \\
\bottomrule
\end{tabular}
\label{decoders}
\end{table}

\vspace{-0.5em}

\subsection{Qualitative Examples}

\begin{figure}[H]
\centering
\includegraphics[height=5cm, keepaspectratio]{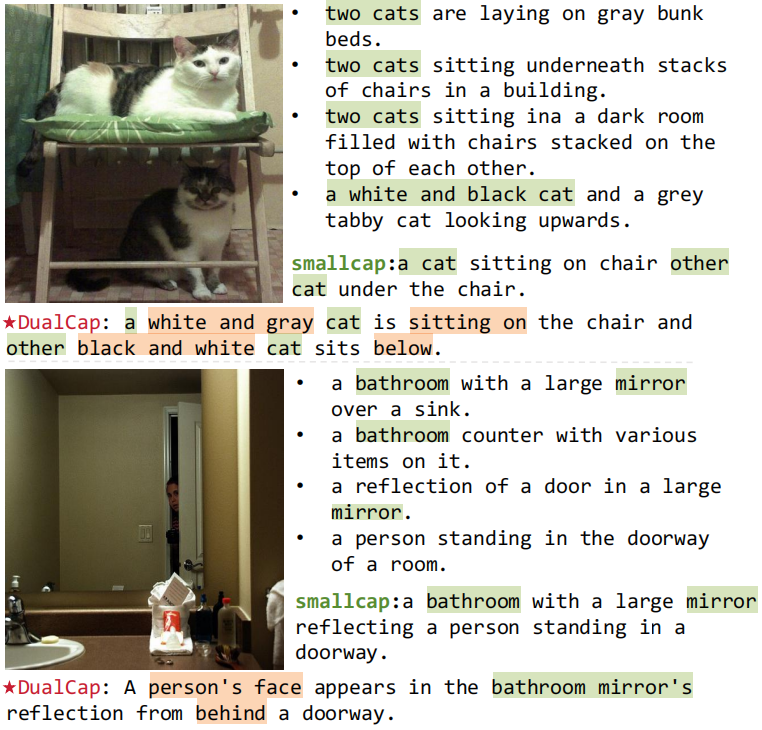} 
\caption{Through using dual retrieval mechanism, DualCap outperforms baselines like SmallCap\cite{ramos2022smallcap} by capturing superior fine-grained detail.}
\Description{Comparison of description generation effects}
\label{fig:qualitative_comp}
\end{figure}

 Figure \ref{fig:qualitative_comp} provides a qualitative comparison that illustrates how DualCap exceeds the baseline model in generating detailed and visually-grounded captions.  In the first example, while the SmallCap baseline produces a factually correct but generic description of two cats, DualCap leverages its Image-to-Image (I2I) retrieval path to capture fine-grained attributes. It correctly identifies the distinct coloration of each animal, describing "a white and grey cat" and its "black and white cat". This result validates that our I2I-sourced visual prompts effectively enrich the model's perception, enabling it to describe specific details beyond the baseline's capability.

This advantage is further highlighted in the second example, which showcases DualCap's robustness in identifying subtle or partially obscured subjects. Faced with a complex bathroom scene where the baseline model overlooks the key subject, DualCap successfully pinpoints the crucial detail: "A person's face appears in the bathroom mirror's reflection from behind a doorway." This comparison suggests how grounding the model in targeted visual cues from the I2I stream helps DualCap avoid generic descriptions and more accurately portray complex visual compositions.


\section{Conclusion}
We introduced DualCap, a lightweight framework that enhances image captioning via a novel dual retrieval mechanism. By sourcing visual prompts from similar scenes to complement standard image-to-text retrieval, our approach achieves significant performance gains across both in-domain and cross-domain datasets with minimal computational overhead. This success validates the potential of dual retrieval for augmented generation systems and opens promising avenues for future work, such as extending the framework to Visual Question Answering.

\bibliographystyle{ACM-Reference-Format}
\bibliography{sample-base}

\appendix

\end{document}